\documentclass{bmvc2k}
\usepackage{amsmath,amssymb,dsfont}
\usepackage{bm}
\usepackage{upgreek}
\usepackage{boldline}
\usepackage{algorithm}
\usepackage{setspace}
\definecolor{ocre}{RGB}{0,0,102}
\usepackage{caption}
\usepackage[font={color=ocre}]{caption}
\usepackage[dvipsnames]{xcolor}
\newcommand{\titlesize}{\fontsize{17.25pt}{17.28pt}\selectfont}

\title{
{\titlesize Deep Association Learning for Unsupervised Video Person Re-identification}
}
\addauthor{Yanbei Chen}{yanbei.chen@qmul.ac.uk}{1}
\addauthor{Xiatian Zhu}{eddy@visionsemantics.com}{2}
\addauthor{Shaogang Gong}{s.gong@qmul.ac.uk}{1}

\addinstitution{
Computer Vision Group, \\
School of Electronic Engineering and Computer Science, \\
Queen Mary University of London, \\ London E1 4NS, UK
}
\addinstitution{Vision Semantics Ltd., \\
London E1 4NS, UK.
}

\runninghead{Y. Chen, X. Zhu, S. Gong}{Deep Association Learning}

\begin{document}

\maketitle
\begin{abstract}
Deep learning methods have started to dominate the
research progress of video-based person re-identification
(re-id). However, existing methods mostly consider supervised learning, which
requires exhaustive manual efforts for labelling cross-view pairwise
data. Therefore, they severely lack scalability and practicality in real-world
video surveillance applications.  
In this work, to address the
video person re-id task, we formulate a novel {\em Deep Association Learning}
(DAL) scheme, the first end-to-end deep learning method 
using none of the identity labels in model initialisation and training. 
DAL learns a deep re-id matching model by jointly optimising two
margin-based association losses in an end-to-end manner, 
which effectively constrains the association of each frame to the best-matched 
intra-camera representation and cross-camera representation.
%Each loss is formulated by constraining association of each frame to the best-matched 
%intra-camera tracklet representation or cross-camera representation.
Existing standard CNNs can be readily employed within our DAL scheme.
Experiment results demonstrate that our proposed DAL significantly outperforms current state-of-the-art
unsupervised video person re-id methods on three benchmarks: PRID 2011, iLIDS-VID and MARS.
\end{abstract}

%-------------------------------------------------------------------------
\section{Introduction}
\label{sec:intro}
Person re-identification (re-id) aims to match persons 
across disjoint camera views distributed at different locations 
\cite{gong2014person}.
While most recent re-id methods rely on static images
\cite{li2014deepreid,ahmed2015improved,xiao2016learning,wang2016joint,li2017person,sun2017svdnet,zheng2017unlabeled,chen2017person,li2018harmonious,zhong2017camera,wang2018person,zhu2017fast},
video-based re-id has gained increasing attention
\cite{hirzer2011person,wang2014person,wang2016person,zhu2016video,zheng2016mars,you2016top,mclaughlin2016recurrent,yan2016person,zheng2016mars,zhou2017see,xu2017jointly}
due to the rich space-time information inherently carried in the video tracklets. 
%In essence, a
A video tracklet is a sequence of images that
captures rich variations of the same person 
in terms of occlusion, background clutter, viewpoint, human poses, etc, 
which can naturally be used as informative data sources for person re-id.
The majority of current techniques in video person re-id consider the supervised 
learning context, which 
imposes a strong assumption on the availability of identity (ID) labels for
every camera pair 
therefore allowing more powerful and discriminative re-id models to be learned 
when given relatively small-sized training data. % from a small number of camera pairs.
% for learning the feature representation and matching
%model optimal for that supervised camera pair.
% 
However, supervised learning methods are weak in scaling %a model
to real-world deployment beyond the labelled training data domains. 
In practice, exhaustive manual annotation at every camera pair 
is not only prohibitively expensive for a large identity population across a large camera network, 
but it is also implausible due to insufficient designated persons reappearing in every camera pair. 
In this regard, unsupervised video re-id is a more realistic task 
that is worth studying to improve the scalability of re-id models in practical use.
%
%This limitation 
%reduces the usability of existing supervised video re-id methods. 

\begin{figure}[!t]
\centering
\includegraphics[width=0.98\textwidth]{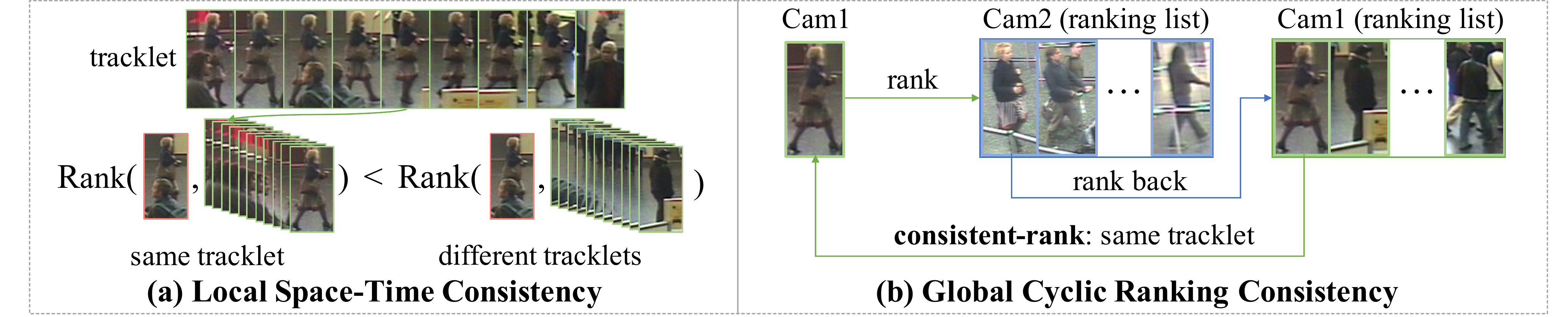}
\vspace{1.0em}
\caption{
Two types of consistency in our Deep Association Learning scheme. 
(a) Local space-time consistency: Most images from the same
tracklet generally depict the same person. 
(b) Global cyclic ranking consistency: Two tracklets from different
cameras are highly associated if they are {\em mutually} the nearest
neighbour returned by a cross-view ranking. %in a cross-view way. % each camera view 
%{\em recursively}.
}
\label{fig:idea}
\vspace{-0.5em}
\end{figure}

Unsupervised learning methods~\cite{ma2017person,liu2017stepwise,ye2017dynamic,liu2015spatio,karanam2015person,wang2018reid} 
are particularly essential 
when the re-id task needs to be performed on a large amount of unlabelled
video surveillance data cumulated continuously over time, whilst the pairwise %per-camera-pair
ID labels cannot be easily acquired for supervised model learning. 
Due to the inherent nature of unsupervised learning, existing methods
suffer from significant performance degradations when compared to 
supervised learning methods in video person re-id. 
For instance, the state-of-the-art rank-1 re-id matching rate on MARS~\cite{zheng2016mars} is only 
36.8\% by unsupervised learning~\cite{ye2017dynamic}, 
as compared to 82.3\% by supervised learning~\cite{li2018diversity}. 
In fact, even the latest video-based unsupervised learning models~\cite{liu2017stepwise,ye2017dynamic} for person re-id 
still lack a principled mechanism
to explore the more powerful representation-learning capabilities of 
deep Convolutional Neural Networks (CNNs)~\cite{bengio2013representation} for jointly learning 
an expressive embedding representation and a discriminative re-id matching model in an end-to-end manner. 
It is indeed not straightforward to formulate a deep learning scheme 
for unsupervised video-based person re-id due to:
(1) The general supervised learning nature of deep CNN networks: 
most deep learning objectives are formulated on labelled training data;
% such as the cross-camera tracklet pairs with ID labels; 
%
(2) The cross-camera variations of the same-ID tracklet pairs from disjoint camera views
and the likelihood of different people being visually similar in public space, 
which collectively render the nearest-neighbour distance measure unreliable 
to capture the cross-view person identity matching for guiding the model learning. 

In this work, we aim to tackle the task of unsupervised video person re-id by an end-to-end optimised deep learning scheme without utilising any ID labels. 
Towards this aim, we formulate a novel {\em unsupervised} \textbf{Deep
  Association Learning} (DAL) scheme designed specifically to explore
  two types of {\em consistency}, including 
%two {\em consistency} constraints intrinsic to video tracklets, including   
%auto-generated video tracklets: % \cite{felzenszwalb2010object,dehghan2015gmmcp}:
(1) {\em local space-time consistency} within each tracklet from the same camera view, 
and (2) {\em global cyclic ranking consistency} between tracklets across disjoint camera views
(Figure \ref{fig:idea}).
%DAL assumes no ID labelled tracklets.
In particular, we define two margin-based association losses, %frame-to-tracklet association losses, 
with one derived from the intra-camera tracklet representation updated incrementally
on account of the {\em local space-time consistency},
and the other derived from the cross-camera representation
learned continuously based on the {\em global cyclic ranking consistency}. 
Importantly, this scheme enables the deep model to start with learning from the local consistency,
whilst incrementally self-discovering more cross-camera highly associated tracklets 
subject to the global consistency for progressively enhancing discriminative feature learning. % in an progressive manner.
%
%In this way, DAL imposes a batch-wise self-supervised learning cycle and eliminates the need for manually
%labelled supervision in model learning.
Overall, our DAL scheme imposes batch-wise self-supervised learning cycles 
to eliminate the need for manual labelled supervision in the course of model training. 

{\bf Our contribution} is three-fold:
{\bf (I)} We propose for the first time an end-to-end deep learning scheme for unsupervised video person re-id without imposing any human knowledge on identity information. 
{\bf (II)} We formulate a novel {\em Deep Association Learning} (DAL) scheme, 
%{\bf (II)} We introduce the idea of {\em Deep Association Learning} (DAL)
with two discriminative association losses derived from  
%with two innovative association losses derived from  
(1) 
{\em local space-time consistency} within each tracklet %in each camera
and (2) 
{\em global cyclic ranking consistency} between tracklets across disjoint camera views.
%
%This scheme is formulated for %enabling the capability of
%concurrently self-discovering the most likely ID association (without
%labelling knowledge) between tracklets
%both within-camera and across-cameras by batch-wise incremental deep learning.
%
Our DAL loss formulation allows typical deep CNNs to be readily 
trained by standard stochastic gradient descent algorithms.
%
%Existing alternative methods lack such an unsupervised tracklet learning mechanism for video re-id.
%
{\bf (III)} Extensive experiments demonstrate the advantages of DAL over the state-of-the-art unsupervised video person re-id methods
on three benchmark datasets: PRID2011 \cite{hirzer2011person}, iLIDS-VID \cite{wang2014person}, and MARS \cite{zheng2016mars}. 

%-------------------------------------------------------------------------
\section{Related Work}
{\bf Unsupervised Video-based Person Re-identification} has started to attract increasing research interest recently~\cite{karanam2015person,khan2016unsupervised,ma2017person,ye2017dynamic,liu2017stepwise}. 
The commonality of most existing methods is to discover the matching correlations between tracklets across cameras. 
%The commonality shared by most existing works on unsupervised video person re-id is to discover the matching correlations between tracklets captured under different cameras. 
For example, Ma et al.~\cite{ma2017person} formulate a time shift dynamic warping model to automatically pair cross-camera tracklets by matching partial segments of each tracklet generated over all time shifts. 
Ye et al.~\cite{ye2017dynamic} propose a dynamic graph matching method to mine the cross-camera labels for iteratively learning a discriminative distance metric model. 
Liu et al.~\cite{liu2017stepwise} develop a stepwise metric learning method to progressively estimate the cross-camera labels; but it requires stringent video filtering to obtain 
one tracklet per ID per camera for discriminative model initialisation. 
The proposed Deep Association Learning (DAL) method in this
work differs significantly from previous works in three aspects:  
(1) Unlike~\cite{ma2017person,liu2017stepwise}, our DAL does not require additional manual effort to select tracklets for model initialisation, which results in better scalability to large-scale video data. 
%therefore more scalable to large-scale video data.
(2) All existing methods rely on a good external feature extractor for metric learning; 
while our DAL jointly learns a re-id matching model with discriminative representation in a fully end-to-end manner. 
(3) Our DAL uniquely utilises the intra-camera  
local space-time consistency and cross-camera global cyclic ranking 
consistency to formulate the learning objective with a relatively low computational cost. 

\vspace{0.1em}
\noindent {\bf Deep Metric Learning} 
aims to learn a nonlinear mapping that transforms input images into a feature representation space, in which 
the distances within the same class are enforced to be small
whilst the distances between different classes are maintained large. 
%the distances between images of the same class are enforced to be small
%whilst the distances between images of different classes are maintained large. 
%It is at the core of building a re-id model.
%Given that learning an embedding representation that properly reflects the similarities between person images, 
A variety of deep distance metric learning methods have been proposed
to solve the person re-id problem~\cite{li2014deepreid,yi2014deep,ahmed2015improved,ding2015deep,liu2016multi,wang2016joint,cheng2016person,chen2016deep,mclaughlin2016recurrent,chen2017beyond,hermans2017defense,xu2017jointly}, 
among which the most popular learning constraint is pairwise comparison~\cite{li2014deepreid,yi2014deep} 
or triplet comparison~\cite{prosserbmvc10,paisitkriangkraicvpr15,hermans2017defense} (also known as relative distance comparison~\cite{zhengpami13,ding2015deep}). 
For pairwise comparison, %in deep learning, 
a binary classification learning objective~\cite{li2014deepreid,ahmed2015improved} or a
Siamese network with a similarity measure
objective~\cite{yi2014deep,mclaughlin2016recurrent,xu2017jointly} is
typically adopted to learn a nonlinear mapping that  
outputs pairwise similarity scores. % for each pair of inputs. 
For triplet comparison, 
% in deep learning, 
a margin-based hinge loss with a batch construction strategy for triplet
generation~\cite{ding2015deep,hermans2017defense} is often deployed to
maximise the relative distance between matched pairs and unmatched
pairs of inputs.
As opposed to most supervised deep metric learning methods in person
re-id, our DAL learns a deep embedding representation in an unsupervised fashion. 
Instead of grounding the learning objective based on pairwise or triple-wise
comparison between a few labelled samples, e.g., three samples as a triplet,
our DAL uniquely learns two set of anchors as the intra-camera and cross-camera tracklet representations, 
which allows to measure the pairwise similarities between each image frame and all
the tracklet representations to formulate the unsupervised learning objectives. %from the same camera view
%without the need of intricate batch construction. 

%-------------------------------------------------------------------------
\vspace{-0.2cm}
\section{Deep Association Learning}

\begin{figure}[!t]
\centering
\includegraphics[width=0.99\textwidth]{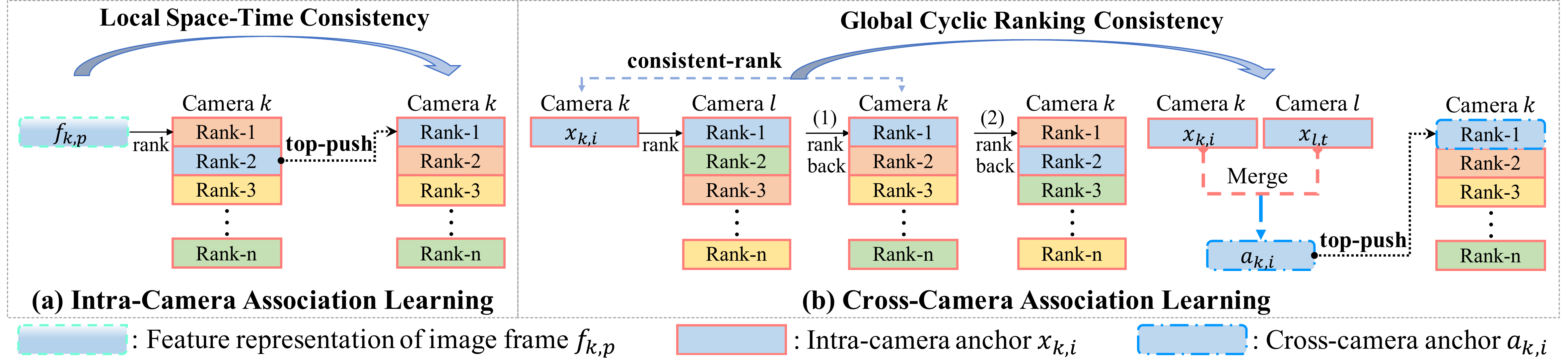}
\vspace{1.0em}
\caption{Illustration of Deep Association Learning:
	(a) Intra-camera association learning based on the local
	space-time consistency within tracklets (Sec.~\ref{intra-cam}). 
	(2) Cross-camera association learning based on the {global cyclic ranking consistency} on cross-camera tracklets (Sec.~\ref{cross-cam}).
	Best viewed in colour.
%(a) Intra-camera association learning derive the top-push margin-based association loss based on the association to the corresponding {\em intra-camera anchors} $x_{k,i}$ learned on account of the {\em temporal consistency} (Sec.~\ref{intra-cam}). 
%(2) Cross-camera association learning derive another top-push margin-based association loss based on the association to the corresponding {\em cross-camera anchors} $a_{k,i}$ learned based on the {\em cyclic ranking consistency} (Sec.~\ref{cross-cam}).
%Different colours correspond to representation of different video tracklets. Best viewed in colour.
}
\label{fig:a_rank}
\vspace{-0.5em}
\end{figure}

\noindent {\bf Approach Overview.} 
Our goal is to learn a re-id matching model to discriminate the
appearance difference and reliably associate the video tracklets
across disjoint camera views without utilising any ID labels. % for training data. 
Towards this goal, we propose a novel {\em Deep Association Learning} (DAL)
scheme that optimises a deep CNN model based on the learning objective %two margin-based association losses 
derived based on two types of consistency. %without any cross-view ID labelled data.  
As illustrated in Figure~\ref{fig:a_rank}, we explore the {\em local
  space-time consistency} and {\em global cyclic ranking consistency}
to formulate two top-push margin-based association losses. 
In particular, two sets of ``anchors'' are gradually learned all along the training process %incrementally learned during training 
for our loss formulation. They are {\bf (1)} a set of {\em intra-camera anchors}
$\{x_{k,i}\}_{i=1}^{N_k}$ that denote the intra-camera feature
representations of $N_k$ tracklets under camera $k$; and {\bf (2)} a
set of {\em cross-camera anchors} $\{a_{k,i}\}_{i=1}^{N_k}$, with each
representing the cross-camera feature representation merged by the
intra-camera feature representations of two highly associated
tracklets from disjoint camera views. 
Overall, the DAL scheme consists of two batch-wise iterative
procedures: {\bf (a)} intra-camera association learning and {\bf (b)}
cross-camera association learning, as elaborated in the following. 

%\vspace{-1em}
\vspace{-0.2cm}
\subsection{Intra-Camera Association Learning}
\label{intra-cam}
%\vspace{-0.3em}

Intra-camera association learning aims at discriminating intra-camera video tracklets. 
To this end, we formulate a top-push margin-based intra-camera association loss 
in the form of the hinge loss based on
the ranking relationship of each image frame in association to all the 
video tracklets from the same camera view. 
This loss is formulated in three steps as follows.

\vspace{0.1em}
\noindent{\bf (1) Learning Intra-Camera Anchors.} 
On account of the {\em local space-time consistency} as depicted Figure~\ref{fig:idea}, each video tracklet can simply be represented as a univocal sequence-level feature representation by utilising certain temporal pooling strategy, such as max-pooling or mean-pooling~\cite{mclaughlin2016recurrent,zheng2016mars}. 
%First, we need to obtain the feature representations for both images frames
%and tracklets for enabling their distance based matching.
%The frame feature vector $f_{k,p}$ can be obtained by
%the up-to-date deep model at any time.
%For tracklets, one common way is by 
%max-pooling or mean-pooling~\cite{mclaughlin2016recurrent,zheng2016mars}.
This, however, is time-consuming to compute at each mini-batch
learning iteration, as it requires to feed-forward all image frames of
each video tracklet through the deep model.
To overcome this problem, 
we propose to represent a tracklet from camera $k$ as an {\em
  intra-camera anchor} $x_{k,i}$, which is the intra-camera tracklet representation
 incrementally updated by the frame representation
$f_{k,p}$ of any constituent image frame from the same source tracklet all through the training process. 
%withoutthe need of extracting features for all tracklet frames. 
Specifically, the exponential moving average (EMA) strategy is adopted to update each anchor $x_{k,i}$ as follows. %all through the training process: 

\begin{equation}
x_{k,i}^{t+1} \leftarrow x_{k,i}^{t} - \eta {\big(}\ell_2(x_{k,i}^{t})-\ell_2({f_{k,p}^{t}}){\big)}, 
\;\; \text{if } i = p
\label{eq:anchor_update}
\end{equation}
where $\eta$ refers to the update rate (set to 0.5), $\ell_2({\cdot})$ is $\ell_2$ normalisation (i.e. $\|\ell_2({\cdot})\|_2 = 1$), 
and $t$ is the mini-batch learning iteration.
As $x_{k,i}$ is initialised as the mean of the frame representations for each tracklet and incrementally updated as Eq.~\eqref{eq:anchor_update}, 
the intra-camera anchor is consistently learned all along with 
the model learning progress to represent each tracklet.
 
\vspace{0.1em}
\noindent{\bf (2) Tracklet Association Ranking.} 
Given the set of incrementally updated {\em intra-camera anchors}
$\{x_{k,i}\}_{i=1}^{N_k}$ for camera $k$, the ranking relationship of
the frame representation $f_{k,p}$ %of an image frame 
in association to all {intra-camera anchors} from the same camera $k$ can be
generated based on pairwise similarity measure. 
%Second, we measure the ranking frame-to-tracklet relationships
%in the intra-camera sense.
We use the $\ell_2$ distance to measure the pairwise similarities between an in-batch frame representation $f_{k,p}$ and all the {intra-camera anchor} $\{x_{k,i}\}_{i=1}^{N_k}$. 
Accordingly, a ranking list is obtained by sorting the pairwise similarities of $f_{k,p}$ 
w.r.t. $\{x_{k,i}\}_{i=1}^{N_k}$, 
with the rank-1 (top-1) {intra-camera
  anchor} having the minimal pairwise distance: 
\begin{equation}
\{D_{p,i} | D_{p,i} = {\big \|} \ell_2(f_{k,p}) - \ell_2(x_{k,i}) {\big \|}_2, \; i \in N_k\} \xrightarrow[]{\text{ranking}}
D_{p,t} = \min_{i\in[1,N_k]} D_{p,i}
\label{eq:ranking_list}
\end{equation}
where $\{D_{p,i}\}_{i=1}^{N_k}$ is the set of pairwise distances between $f_{k,p}$ and $\{x_{k,i}\}_{i=1}^{N_k}$; 
while $D_{p,t}$ denotes the pairwise distance between $f_{k,p}$ and the rank-1 tracklet $x_{k,t}$.
%where $\{D_{p,i}\}_{i=1}^{N_k}$ is the set of pairwise distances between the feature representation $f_{k,p}$ and the set of {\em intra-camera anchors} $\{x_{k,i}\}_{i=1}^{N_k}$ under camera $k$. $D_{p,t}$ denotes the top-rank pairwise distance between $f_{k,p}$ and $x_{k,t}$, i.e. $x_{k,t}$ is ranked as rank-1 among $\{x_{k,i}\}_{i=1}^{N_k}$.

\vspace{0.1em}
%\noindent{\bf (3) Intra-Camera Tracklet Association Loss.} 
\noindent{\bf (3) Intra-Camera Association Loss.} 
%For formulating the intra-camera frame-to-tracklet association loss, 
Given the ranking list for the frame representation $f_{k,p}$ (Eq. \eqref{eq:ranking_list}), 
the intra-camera rank-1 tracklet $x_{k,t}$ should 
ideally correspond to the source tracklet $x_{k,p}$ that contains the same constituent frame 
due to the {\em local space-time consistency}. 
We therefore define a top-push margin-based intra-camera association loss %, in the form of hinge loss,  
to enforce proper association of each frame to the source tracklet for discriminative model learning: 
%
%following the {\em temporal consistency}, each image frame is naturally most associated to the same video tracklet that the image belongs to. Hence, to achieve learning discrimination on the appearance difference among different video tracklets, a top-push margin-based intra-camera association loss is formulated to ensure the feature representation $f_{k,p}$ of an image frame is assigned as the top-rank to the corresponding {\em intra-camera anchors} $x_{k,p}$ that denotes the same video tracklet. 
%In particular, the intra-camera association loss is derived in the form of the hinge loss: 
%\vspace{-0.5em}
\begin{equation}
%\vspace{-0.5em}
\mathcal{L}_I=\left\{\begin{aligned}
&{[D_{p,p} - D_{p,t} +m]}_{+}, &\text{if } p \neq t 
\;\; & \text{(The rank-1 is not the source tracklet)}
\\
&{[D_{p,p} - {\overline{D_{j,t}} +m]}}_{+}, &\text{if } p = t
\;\; & \text{(The rank-1 is the source tracklet)} \\
\end{aligned} \right.
\label{eq:hardpush}
\end{equation}
where $[\cdot]_{+}\; = \;\max(0,\;\cdot)$,
$D_{p,p}$ is the pairwise distance between $f_{k,p}$ and $x_{k,p}$ (the source tracklet), 
$\overline{D_{j,t}} = \frac{1}{M}\sum_{j=1}^M{D_{j,t}}$ is the averaged rank-1 pairwise distance of the $M$ sampled image frames from camera $k$
in a mini-batch. 
$m$ is the margin that enforces the deep model to assign the source tracklet as the top-rank. 
%and $m$ is the discrimination margin 
%between the source and non-source tracklets
%against a frame. 
%
%Moreover, $\mathcal{L}_I$ is derived based on one association condition: whether the top-rank is true or false, i.e. whether {\em intra-camera anchor} corresponds to the same tracklet. 
More specifically, if the rank-1 is not the source tracklet (i.e. $p \neq t$), 
$\mathcal{L}_I$ will correct the model by imposing a large penalty
to push the source tracklet to the top-rank.  
Otherwise, $\mathcal{L}_I$ will further minimise the intra-tracklet variation w.r.t. the averaged rank-1 pairwise distance in each mini-batch. 
Since $\mathcal{L}_I$ is computed based on 
the sampled image frames and the up-to-date {intra-camera anchors} in each mini-batch, 
it can be efficiently optimised by the standard stochastic gradient descent to adjust the deep CNN parameters iteratively. %seamlessly integrated into the standard deep model optimisation. 
Overall, $\mathcal{L}_I$ encourages to learn the discrimination on intra-camera tracklets for facilitating the 
more challenging cross-camera association, as described next.
 
\vspace{-0.2cm}
\subsection{Cross-Camera Association Learning}
\label{cross-cam}
%A key of video re-id model learning is to impose the 
%cross-camera ID pairing information of tracklets, which is not provided %does not exist
%in unsupervised learning. 
A key of video re-id is to leverage the 
cross-camera ID pairing information for model learning.
However, such information is missing in unsupervised learning. 
We overcome this problem by self-discovering the cross-camera tracklet association in a progressive way 
during model training. 
To permit learning expressive representation invariant to 
the cross-camera appearance variations inherently carried in 
associated tracklet pairs from disjoint camera views, 
we formulate another top-push margin-based intra-camera association loss 
in the same form as Eq.~\eqref{eq:hardpush}. 
%We adopt the same loss formulation (Eq. \eqref{eq:hardpush}) as $\mathcal{L}_I$, 
Crucially, we extend the tracklet
representation to carry the information of cross-camera appearance variations 
by incrementally learning a set of {\em cross-camera anchors}. 
%to carry the information of cross-camera variations. 
%called {\em cross-camera tracklet anchors}.
% 
%Specially, to minimise incorrect cross-camera tracklet association
%at the absence of ID labels in model learning,
%%
%we explore the {\em global cyclic ranking consistency} to more reliably associate tracklets across disjoint camera views. 
%
This intra-camera association loss is formulated in three steps as below.

\vspace{0.1em}
\noindent{\bf (1) Cyclic Ranking.} 
Given the incrementally updated {intra-camera anchors} (Eq.~\eqref{eq:anchor_update}), 
we propose to exploit the underlying relations between tracklets for  
discovering the association between tracklets across different cameras.
%we propose to exploit the underlying landscape of feature distribution on tracklets %under each camera 
%for discovering reliable cross-camera association. 
Specifically, a cyclic ranking process is conducted to attain the pair of %discover 
highly associated {intra-camera anchors} across cameras as follows.  
%all the {\em intra-camera anchors} together reflect the global landscape of all the tracklets in the feature space. As shown in Fig.~\ref{fig:idea}(b), \ref{fig:a_rank}, to discover the associations between tracklets across different cameras during training, we utilise a cyclic ranking process to find the highly associated {\em intra-camera anchors} from two non-overlapping camera views: 
\begin{equation}
\begin{aligned}
{x_{k,i}
\xrightarrow[]{\text{ranking in cam $l$}}
Dc_{p,t} = min_{i\in[1,N_l]} 
\xrightarrow[]{\text{ranking back in cam $k$}}
Dc_{q,j} = min_{i\in[1,N_k]}}
\\
\includegraphics[width=0.86\textwidth]{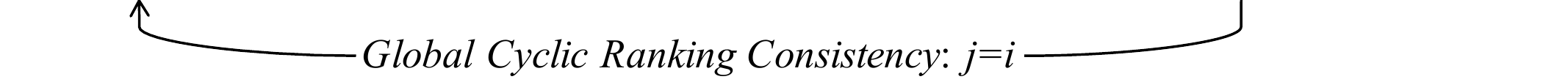}
\end{aligned}
\label{eq:cyclic}
%\vspace{-0.1em}
\end{equation}
where $Dc_{p,t}$ denotes the cross-camera pairwise distance between two {intra-camera anchors}: $x_{k,p}$ from camera $k$ and $x_{l,t}$ from another camera $l$. 
Both $Dc_{p,t}$ and $Dc_{q,j}$ denote the rank-1 pairwise distance. 
The pairwise distance and the ranking are computed same as Eq.~\eqref{eq:ranking_list}. 
With Eq.~\eqref{eq:cyclic}, we aim to discover the most associated intra-camera anchors across cameras
%tracklets by their {intra-camera anchors} 
under the criterion of {\em global cyclic ranking consistency}:  
$x_{k,p}$ and $x_{l,t}$ are mutually the rank-1 match pair to each other 
when one is given as a query to search for the best-matched {intra-camera
anchor} in the other camera view. %, and vice versa recursively. 
This cyclic ranking process is conceptually related to
 the cycle-consistency constraints formulated to enforce the %in utilising 
%mutual consistency as regularise the model learning
pairwise correspondence between similar instances \cite{zhou2016learning,sener2016learning,godard2017unsupervised}. 
In particular, our {\em global cyclic ranking consistency} in this process 
%is particularly utilised to progressively discover the highly associated tracklets 
%across disjoint camera views during training. 
aims to exploit the mutual consistency induced by transitivity for discovering the highly associated tracklets across disjoint camera views all along the model training process. % by a cyclic ranking process. 

\vspace{0.1em}
\noindent{\bf (2) Learning Cross-Camera Anchors.} 
%Once finding a cycle-consistent match for a tracklet $x_{k,p}$,
Based on {\em global cyclic ranking consistency}, we define the cross-camera representation as a {\em cross-camera anchor} $a_{k,i}$ by merging two highly associated {intra-camera anchors} as depicted in Figure~\ref{fig:a_rank} and detailed below.
%to denote the cross-view feature representations that are updated as the arithmetic mean of any two highly associated {\em intra-camera anchors} conditioned on the {\em cyclic ranking consistency}: 
%\vspace{-0.5em}
\begin{equation}
%\vspace{-0.5em}
a_{k,i}^{t+1} \leftarrow
\left\{\begin{aligned}
&\frac{1}{2}{\Big(}\ell_2(x_{k,i}^{t+1})+\ell_2(x_{l,t}^{t}){\Big)},&\text{if } j = i &\quad \text{(Cyclic ranking consistent)} \\
&x_{k,i}^{t+1},&\text{others} \\
\end{aligned} \right.
\label{eq:a_anchor_update}
\end{equation}
where $a_{k,i}$ is simply a counterpart of $x_{k,i}$.
%; but it is likely to denote the cross-camera representation 
%conditioned on the {\em global cyclic ranking consistency}. 
%the cross-camera and intra-camera anchor of the same tracklet. 
Each {cross-camera anchor} is updated as the arithmetic mean of two {intra-camera anchors} if the consistency condition is fulfilled (i.e. $j=i$),  
%if a cycle-consistent cross-camera match can be found,
otherwise as the same {intra-camera anchor}. 
As the deep model is updated continuously to
discriminate the appearance difference among tracklets, 
more intra-camera anchors are progressively discovered to be highly associated. %across disjoint camera views. 
%it also progressively discovers more highly associated tracklets across disjoint camera views. 
That is, all along the training process, more {cross-camera anchors} are gradually updated by 
merging the highly associated {intra-camera anchors} 
to carry the information of cross-camera appearance variations induced by the tracklet pairs
that come from disjoint camera views but potentially depict the same identities.

\vspace{0.1em}
%\noindent{\bf (3) Cross-Camera Tracklet Association Loss.} 
\noindent{\bf (3) Cross-Camera Association Loss.} 
Given the continuously updated {\em cross-camera anchors}
$\{a_{k,i}\}_{i=1}^{N_k}$, 
we define another top-push margin-based cross-camera association loss 
in the same form as Eq.~\eqref{eq:hardpush} 
to enable learning from cross-camera appearance variations:  
\begin{equation}
\mathcal{L}_C=\left\{\begin{aligned}
&{[Da_{p,p} - D_{p,t} +m]}_{+}, &\text{if } p \neq t 
\;\; & \text{(The rank-1 is not the source tracklet)}
\\
&{[Da_{p,p} - {\overline{D_{j,t}} +m]}}_{+}, &\text{if } p = t 
\;\; & \text{(The rank-1 is the source tracklet)}
\\
\end{aligned} \right.
\label{eq:a_hardpush}
\end{equation}
where $Da_{p,p}$ denotes the pairwise distance between the frame representation $f_{k,p}$ and the {cross-camera anchor} $a_{k,p}$. 
%of the source tracklet.
%in comparison to the {\em intra-camera anchor} $x_{k,p}$ in the intra-camera association loss  $\mathcal{L}_I$. 
Both $D_{p,t}$ and ${\overline{D_{j,t}}}$ are the same quantities as $\mathcal{L}_I$ in Eq.~\eqref{eq:hardpush}. As depicted in Figure~\ref{fig:a_rank} and in the same spirit as $\mathcal{L}_I$, the cross-camera association loss $\mathcal{L}_C$ enforces the deep model to push the best-associated {cross-camera anchor} as the top-rank, so as to align the frame representation $f_{k,p}$ towards the corresponding cross-camera representation. %minimise the cross-camera variations. 

\vspace{-0.2cm}
\subsection{Model Training}
\noindent{\bf Overall Learning Objective.}
The final learning objective for DAL is to jointly optimise two association losses (Eq.~\eqref{eq:hardpush},
\eqref{eq:a_hardpush}) as follows. 
\begin{equation}
\mathcal{L}_{DAL}=\mathcal{L}_I+\lambda \mathcal{L}_C
\label{eq:final_loss}
\end{equation}
where $\lambda$ is a tradeoff parameter that is set to 1 to 
ensure both loss terms contribute equally to the learning process. 
The margin $m$ in both Eq.~\eqref{eq:hardpush} and Eq. \eqref{eq:a_hardpush} 
is empirically set to 0.2 in our experiments.
The algorithmic overview of model training is summarised in Algorithm \ref{Algorithm}. 
Our implementation is available at: \href{https://github.com/yanbeic/Deep-Association-Learning}{\color{VioletRed}https://github.com/yanbeic/Deep-Association-Learning}.

\vspace{0.1em}
\noindent{\bf Complexity Analysis.}
We analyse the per-batch per-sample complexity cost induced by DAL.
In association ranking (Eq.~\eqref{eq:ranking_list}), the pairwise distances are computed between each in-batch image frame and $N_k$ {intra-camera anchors} for each camera, which leads to a computation complexity of 
$\mathcal{O}{\big (}N_k{\big )}$ for distance computation
and $\mathcal{O}{\big (}N_klog(N_k){\big )}$ for ranking. % respectively. 
Similarly, in cyclic ranking (Eq.~\eqref{eq:cyclic}), the total 
computation complexity is $\mathcal{O}{\big (}N_l+N_k{\big
  )}+\mathcal{O}{\big (}N_llog(N_l)+N_klog(N_k){\big )}$. 
All the distance measures are simply computed 
by matrix manipulation on GPU with single floating point precision 
for computational efficiency.%, which is hence efficient. 
\vspace{-0.2cm}
\begin{algorithm}[h]
	\caption{\normalsize Deep Association Learning.} \label{Algorithm}
	{
	\textbf{Input:} Unlabelled video tracklets captured from different cameras. \\
	\textbf{Output:} A deep CNN model for re-id matching. \\
	\textbf{for} $t=1$ \textbf{to}  \textsl{max\_iter} \textbf{do} \\ 
		\hphantom{~~~~~~}  
		Randomly sample a mini-batch of image frames. \\
		\hphantom{~~~~~~} 
		Network forward propagation.\\
		\hphantom{~~~~~~} 
		Tracklet association ranking on the {\em intra-camera anchors} (Eq.~\eqref{eq:ranking_list}). \\
		\hphantom{~~~~~~} 
		Compute two margin-based association loss terms (Eq.~\eqref{eq:hardpush}, \eqref{eq:a_hardpush}). \\
		\hphantom{~~~~~~} 
		Update the corresponding {\em intra-camera anchors} based on the EMA strategy  (Eq.~\eqref{eq:anchor_update}). \\
		\hphantom{~~~~~~} 
		Update the corresponding {\em cross-camera anchors} based on cyclic ranking (Eq.~\eqref{eq:cyclic}, \eqref{eq:a_anchor_update}). \\
		\hphantom{~~~~~~} 
		Network update by back-propagation (Eq.~\eqref{eq:final_loss}). \\
	\textbf{end for}} %\\
\end{algorithm}

%-------------------------------------------------------------------------
\vspace{-0.5cm}
\section{Experiments}
\vspace{-0.1cm}
\subsection{Evaluation on Unsupervised Video Person Re-ID}
\begin{figure}[h]
	\setlength{\tabcolsep}{0.22em}
	\begin{tabular}{ccc}
		\vspace{-0.3em}
		\bmvaHangBox{\includegraphics[width=4.15cm]{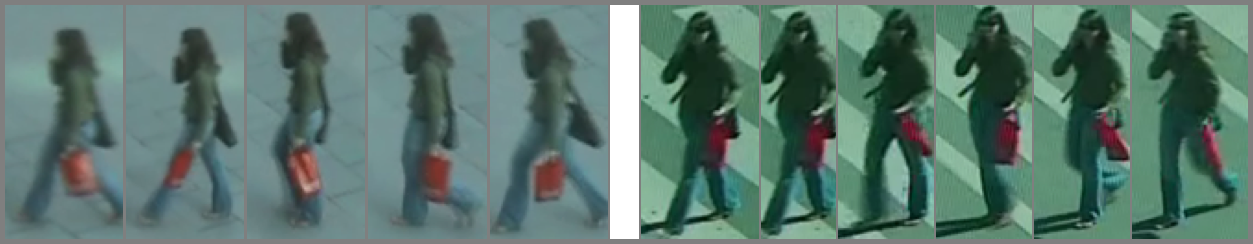}}&
		\bmvaHangBox{\includegraphics[width=4.15cm]{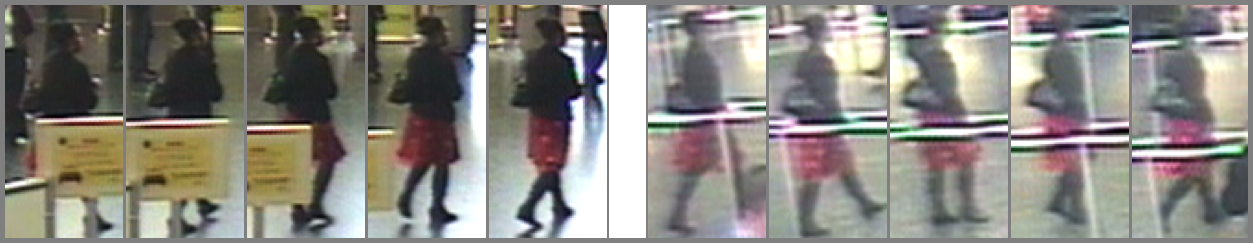}}&
		\bmvaHangBox{\includegraphics[width=4.15cm]{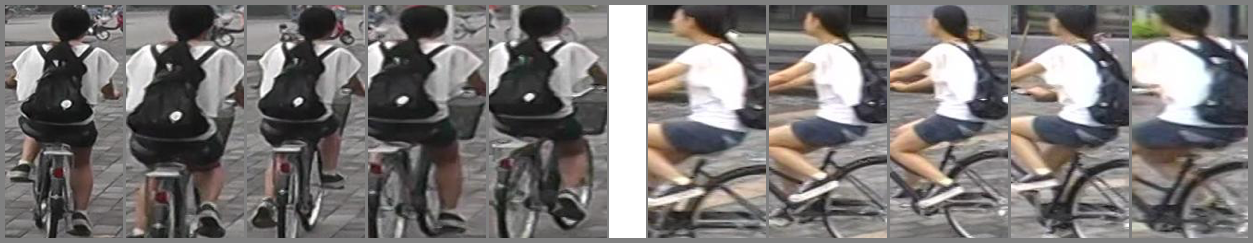}}\\
		\footnotesize{(a) PRID 2011}&\footnotesize{(b) iLIDS-VID}&\footnotesize{(c) MARS}
	\end{tabular}
	\vspace{0.5em}
	\caption{Example pairs of tracklets from three benchmark datasets. Cross-camera variations include changes in illumination, viewpoints, resolution, occlusion, background clutter, human poses, etc.
	}
	\label{fig:dataset}
	%\vspace{-0.8em}
\end{figure}

\noindent {\bf Datasets.}
We conduct extensive experiments on three video person re-id benchmark datasets, including PRID 2011~\cite{hirzer2011person}, iLIDS-VID~\cite{wang2014person} and MARS~\cite{zheng2016mars} (Figure~\ref{fig:dataset}). The PRID 2011 dataset contains 1,134 tracklets captured from two disjoint surveillance cameras with 385
and 749 tracklets from the first and second cameras. Among all video tracklets, 200 persons are captured in both cameras. The iLIDS-VID dataset includes 600 video tracklets of 300 persons. Each person has 2 tracklets from two non-overlapping camera views in an airport arrival hall. 
The MARS has a total of 20,478 tracklets of 1,261 persons captured from a camera network with 6 near-synchronized cameras at a university campus. All the tracklets were automatically generated by the DPM detector \cite{felzenszwalb2010object} and the GMMCP tracker \cite{dehghan2015gmmcp}. 
%The MARS is a large-scale video-based person re-id dataset, with a total of 20,478 tracklets of 1,261 persons captured from a camera network with 6 near-synchronized cameras placed at the campus. All the tracklets are automatically generated by the DPM detector and the GMMCP tracker. 

\vspace{0.1em}
\noindent {\bf Evaluation Protocols.}
For PRID 2011, following \cite{wang2014person,ye2017dynamic,liu2017stepwise} we use the tracklet pairs from 178 persons, with each tracklet containing over 27 frames. These 178 persons are further randomly divided into two halves (89/89) for training and testing. 
For iLIDS-VID, all 300 persons are also divided into two halves (150/150) for training and testing. 
For both datasets, we repeat 10 random training/testing ID splits as \cite{wang2014person} to ensure statistically stable results. The average Cumulated Matching Characteristics (CMC) are adopted as the performance metrics. 
For MARS, we follow the standard training/testing split~\cite{zheng2016mars}: all tracklets of 625 persons for training and the remaining tracklets of 636 persons for testing. 
Both the averaged CMC and the mean Average Precision (mAP) are used to measure re-id performance on MARS.
Note, our method does not utilise any ID labels for model initialisation or training.

\vspace{0.1em}
\noindent {\bf Implementation Details.}
We implement our DAL scheme in Tensorflow~\cite{abadi2016tensorflow}. 
To evaluate its generalisation ability of incorporating with different network architectures, 
we adopt two standard CNNs    
as the backbone networks: ResNet50~\cite{he2016deep} and MobileNet~\cite{howard2017mobilenets}. 
Both deep models are initialised with weights pre-trained on ImageNet~\cite{deng2009imagenet}.
On the small-scale datasets (PRID 2011 and iLIDS-VID), we apply the RMSProp optimiser~\cite{tieleman2012lecture} to train the DAL for $2{\times}10^4$ iterations, with an initial learning rate of 0.045 and decayed exponentially by 0.94 every 2 epochs.
On the large-scale dataset (MARS), we adopt the standard stochastic gradient descent (SGD) to train the DAL for $1{\times}10^5$ iterations, with an initial learning rate of 0.01 and decayed to 0.001 in the last $5{\times}10^4$ iterations. 
The batch size is all set to 64. 
At test time, 
we obtain the tracklet representation
by max-pooling on the image frame features
followed by
$\ell_2$ normalisation.
We compute the $\ell_2$-distance between the cross-camera tracklet representations as the similarity measure for the final video re-id matching.

\begin{table}[!t]
	\centering
	\footnotesize
	\setlength{\tabcolsep}{0.5em}
	\begin{tabular}{l|cccc|cccc|cccc|c}
		\hlineB{2}
		Datasets & \multicolumn{4}{c|}{PRID 2011} & \multicolumn{4}{c|}{iLIDS-VID} & \multicolumn{5}{c}{MARS} \\ \hline
		Rank@$k$ & 1 & 5 & 10 & 20 & 1 & 5 & 10 & 20 & 1 & 5 & 10 & 20 & mAP \\ \hline\hline
		DVDL~\cite{karanam2015person} & 40.6 & 69.7 & 77.8 & 85.6  & 25.9 & 48.2 & 57.3 & 68.9 & - & - & - & - & - \\ 
		STFV3D~\cite{liu2015spatio} & 42.1 & 71.9 & 84.4 & 91.6 & 37.0 & 64.3 & 77.0 & 86.9 & - & - & - & - & -  \\
		MDTS-DTW~\cite{ma2017person} & 41.7 & 67.1 & 79.4 & 90.1 & 31.5 & 62.1 & 72.8 & 82.4 & - & - & - & - & - \\ 
		UnKISS~\cite{khan2016unsupervised} & 59.2 & 81.7 & 90.6 & 96.1 & 38.2 & 65.7 & 75.9 & 84.1 & - & - & - & - & - \\
		DGM+IDE~\cite{ye2017dynamic} & 56.4 & 81.3 & 88.0 & 96.4 & 36.2 & 62.8 & 73.6 & 82.7 & 36.8 & 54.0 & 61.6 & 68.5 & 21.3 \\
		Stepwise~\cite{liu2017stepwise} & 80.9 & 95.6 & {\color{red} 98.8} & {\color{blue} 99.4} & 41.7 & 66.3 & 74.1 & 80.7 & 23.6 & 35.8 & - & 44.9 & 10.5 \\
		\hline
		{\bf DAL (ResNet50)} &  
		{\color{red} 85.3} & {\color{red} 97.0} & {\color{red} 98.8} & {\color{red} 99.6} & 
		{\color{red} 56.9}  & {\color{red} 80.6}  & {\color{red} 87.3}  & {\color{red} 91.9}  & 
		{\color{blue} 46.8} & {\color{blue} 63.9} & {\color{blue} 71.6} & {\color{blue} 77.5} & {\color{blue} 21.4}
		\\
		{\bf DAL (MobileNet)} &  
		%{\color{red} 86.1} & {\color{blue} 96.9} & {\color{blue} 98.3} & 99.2 & 
		{\color{blue} 84.6} & {\color{blue} 96.3} & {\color{blue} 98.4} & 99.1 & 
		{\color{blue} 52.8} & {\color{blue} 76.7} & {\color{blue} 83.4} & {\color{blue} 91.6} &
		{\color{red} 49.3} & {\color{red} 65.9} & {\color{red} 72.2} & {\color{red} 77.9} & {\color{red} 23.0} 
		\\
		\hlineB{2}
	\end{tabular}
	\vspace{1.0em}
	\caption{Evaluation on three benchmarks in comparison to the
          state-of-the-art unsupervised video re-id
          methods. {\color{red} Red}: the best
          performance. {\color{blue} Blue}: the second best
          performance. `-': no reported results.} 
	\vspace{-1.em}
	\label{tab:SOTA}
\end{table}

\vspace{0.1em}
\noindent {\bf Comparison to the state-of-the-art methods.} 
We compare DAL against six state-of-the-art video-based unsupervised
re-id methods: DVDL~\cite{karanam2015person},
STFV3D~\cite{liu2015spatio}, MDTS-DTW~\cite{ma2017person},
UnKISS~\cite{khan2016unsupervised}, DGM+IDE~\cite{ye2017dynamic}, and
Stepwise~\cite{liu2017stepwise}.  
Among all methods, DAL is the only unsupervised deep re-id model that is optimised in an end-to-end manner.
%allowing for fully unsupervised end-to-end model optimisation. 
%
Table~\ref{tab:SOTA} shows a clear performance superiority of DAL
over all other competitors on the three benchmark datasets.
In particular, the rank-1 matching accuracy is improved by 4.4\%(85.3-80.9) on PRID 2011, 15.2\%(56.9-41.7) on iLIDS-VID and 12.5\%(49.3-36.8) on MARS. 
This consistently shows the advantage of DAL
over existing methods for unsupervised video re-id
due to the joint effect of optimising two association losses to 
enable learning feature representation 
invariant to cross-camera appearance variations whilst discriminative to appearance difference. 
%exploiting {\em local space-time consistency} and 
%{\em global cyclic ranking consistency} 
%in an end-to-end manner.
% 
Note, 
the strongest existing model DGM+IDE~\cite{ye2017dynamic} additionally
uses ID label information from one camera view for model initialisation, 
whilst Stepwise~\cite{liu2017stepwise} assumes one tracklet per ID per
camera by implicitly using ID labels.
In contrast, DAL uses neither of
such additional label information for model initialisation or training. 
%
%In contrast, the proposed DAL has no any of such assumptions and therefore more scalable to real-world deployments than DGM and Stepwise.
%
%Besides the advantage on model performance, 
%DAL is also superior in terms of the manual efforts involved during model training. DAL is fully unsupervised without any additional need of preselection on tracklets with identity information, as opposed to its best competitors Stepwise~\cite{liu2017stepwise}, DGM+IDE~\cite{ye2017dynamic} that are indeed partially unsupervised due to the requirement of preselecting tracklets for model initialisation.
%The overall best performance and the second best performance are both achieved by DAL using either MobileNet or ResNet50 as the backbone model.
%
More crucially, DAL consistently produces similar strong re-id performance with
different network architectures (ResNet50 and MobileNet), 
which demonstrates its applicability to existing standard CNNs.
%its adaptability to existing CNN models.
%
%This suggests the good generalisation of the proposed DAL loss formulation in integrating standard CNN models.
% 

%-------------------------------------------------------------------------
\vspace{-0.2cm}
\subsection{Component Analyses and Further Discussions}
%\vspace{-0.3em}
\noindent {\bf Effectiveness of two association losses.}
The DAL trains the deep CNN model based on the joint effect of two association losses: 
(1) intra-camera association loss $\mathcal{L}_I$ (Eq.~\eqref{eq:hardpush}) and 
(2) cross-camera association loss $\mathcal{L}_C$ (Eq.~\eqref{eq:hardpush}).
We evaluate the individual effect of each loss term by eliminating the other term 
from the overall learning objective (Eq.~\eqref{eq:final_loss}). 
As shown in Table~\ref{tab:ablation}, jointly optimising two losses leads
to the best model performance. 
This indicates the complementary benefits of the two loss terms in discriminative feature learning.
Moreover, applying $\mathcal{L}_C$ alone has already achieved better 
performance as compared to the state-of-the-art methods in Table \ref{tab:SOTA}.
When comparing with $\mathcal{L}_I{+}\mathcal{L}_C$, applying $\mathcal{L}_C$ alone
only drop the rank-1 accuracy by 3.0\%(84.6-81.6), 5.4\%(52.8-47.4),
1.2\%(49.3-48.1) on PRID 2011, iLIDS-VID, MARS respectively. This
shows that even optimising the cross-camera association loss {\em alone} 
can still yield competitive re-id performance, 
which owes to its additional effect in enhancing cross-camera invariant representation learning 
%enforcing representation learning %more expressive representation 
%invariant to cross-camera appearance variations 
by reliably associating tracklets across disjoint camera views
all along the training process. 

\begin{table}[b]
	\vspace{-1em}
	\centering
	\footnotesize
	\setlength{\tabcolsep}{0.5em}
	\begin{tabular}{l|cccc|cccc|cccc|c}
		\hlineB{2}
		Datasets & \multicolumn{4}{c|}{PRID 2011} & \multicolumn{4}{c|}{iLIDS-VID} & \multicolumn{5}{c}{MARS} \\
		\hline
		Rank@$k$ & 1 & 5 & 10 & 20 & 1 & 5 & 10 & 20 & 1 & 5 & 10 & 20 & mAP \\ \hline\hline
		%DAL (w $\mathcal{L}_I$) 
		$\mathcal{L}_I$ Only
		& 
		62.7 & 85.7 & 92.1 & 96.7 & 
		31.7 & 55.2 & 67.5 & 78.6 & 
		41.6 & 59.0 & 66.2 & 73.2 & 16.8
		\\
		%DAL (w $\mathcal{L}_C$) 
		$\mathcal{L}_C$ Only & 
		81.6 & 95.2 & 98.1 & {\color{red} 99.7} & 
		47.4 & 72.6 & 81.5 & 89.2 & 
		48.1 & 65.3 & 71.4 & 77.6 & 22.6
		\\
		%DAL (w $\mathcal{L}_{I}{+}\mathcal{L}_{C}$) 
		$\mathcal{L}_{I} {+} \mathcal{L}_{C}$
		& 
		{\color{red} 84.6} & {\color{red} 96.3} & {\color{red} 98.4} & 99.1 & 
		%{\color{red} 86.1} & {\color{red} 96.9} & {\color{red} 98.3} & {\color{red} 99.2} & 
		{\color{red} 52.8} & {\color{red} 76.7} & {\color{red} 83.4} & {\color{red} 91.6} & 
		{\color{red} 49.3} & {\color{red} 65.9} & {\color{red} 72.2} & {\color{red} 77.9} & {\color{red} 23.0} 
		\\
		%ID-Supervised & 
		%84.3 & 98.1 & 99.2 & 99.8 & 
		%51.5 & 76.0 & 83.8 & 89.9 & 
		%71.8 & 86.8 & 90.7 & 93.3 & 51.5 \\
		\hlineB{2}
	\end{tabular}
	\vspace{1.em}
	\caption{Effectiveness of two association losses. {\color{red} Red}: the best performance. CNN: MobileNet.}
	\label{tab:ablation}
	%\vspace{-1em}
\end{table}

\vspace{0.1em}
\noindent {\bf Evolution of cross-camera tracklet association.} 
As aforementioned, learning representation robust to cross-camera variations %tracklet association 
is a key to learning an effective video re-id model.
%This hence delivers highly interpretable justification
%for a model performance.
%
\begin{figure}[!t]
	\setlength{\tabcolsep}{0.22em}
	\centering
	\begin{tabular}{ccc}
		\bmvaHangBox{\includegraphics[width=0.5\textwidth]{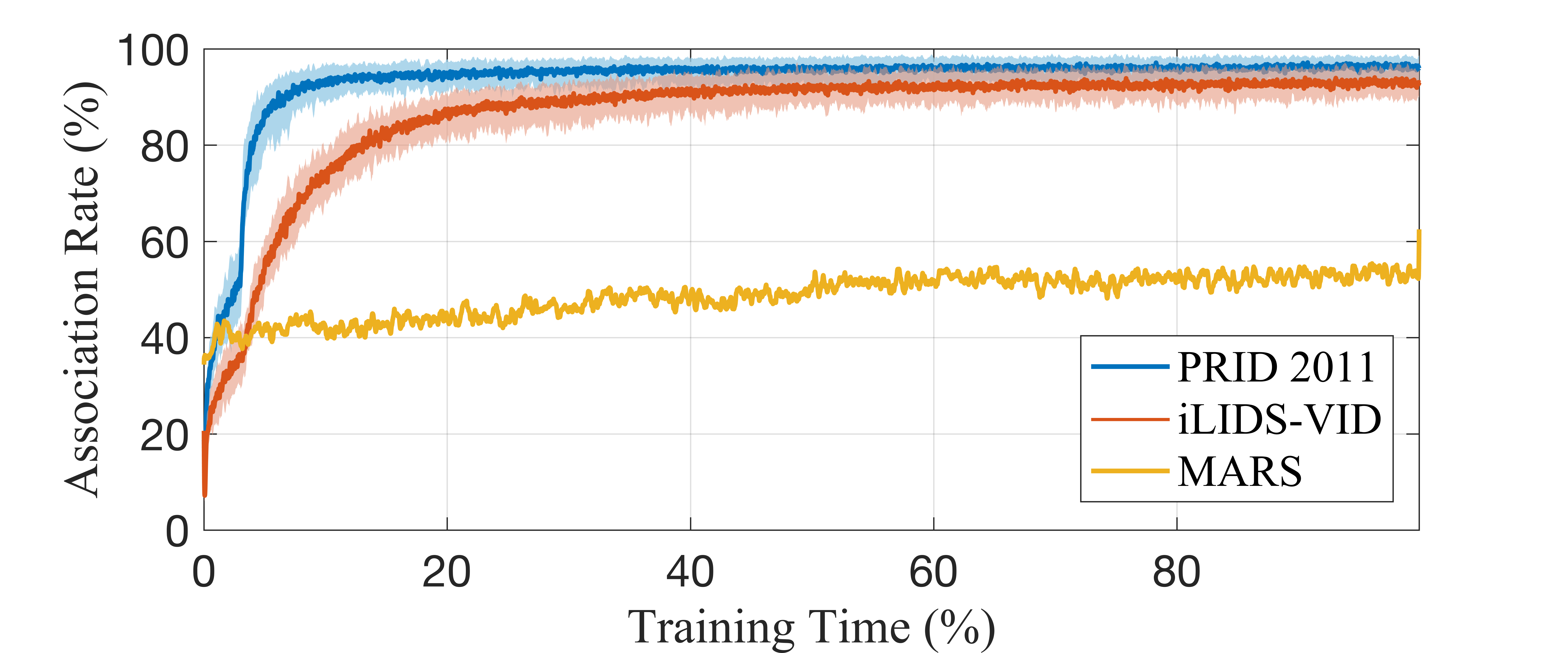}} &
		\bmvaHangBox{\includegraphics[width=0.5\textwidth]{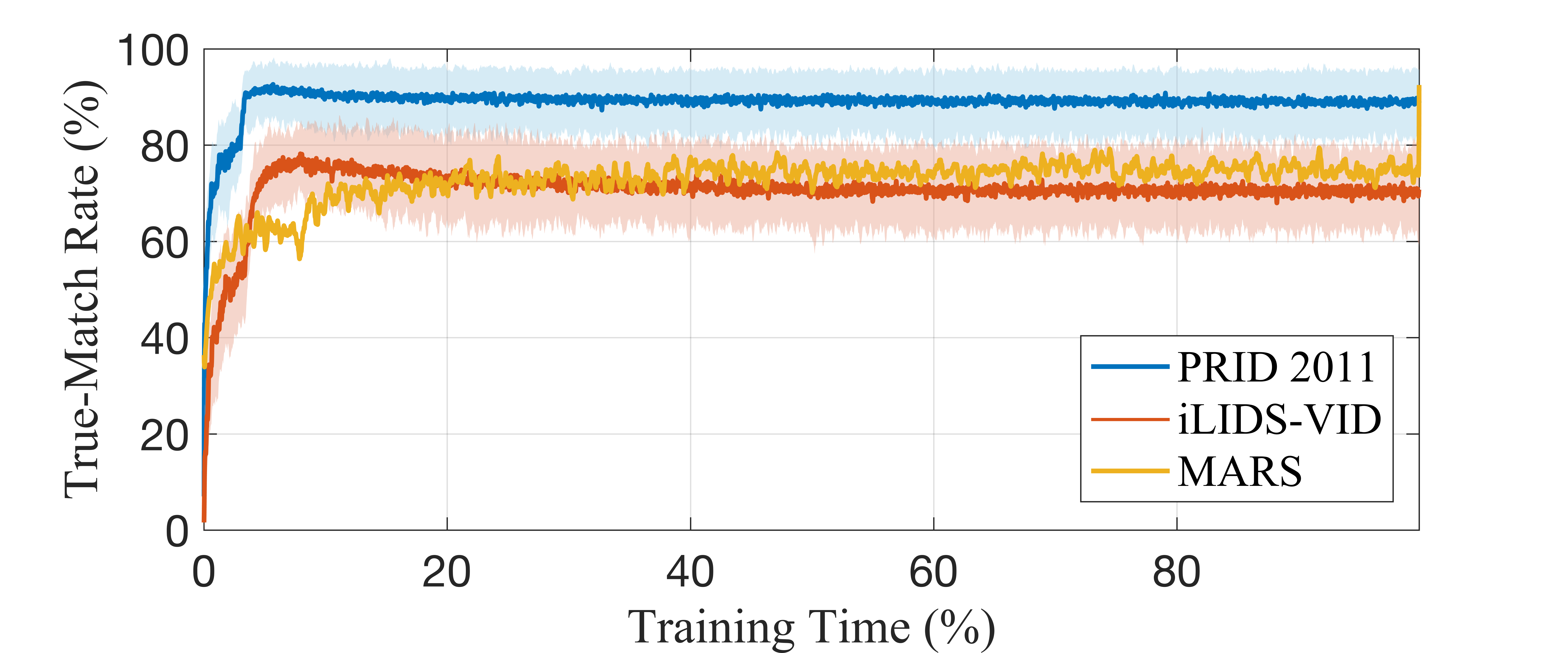}}
		\\
		\small{(a) Evolution on association rate.} &
		\small{(b) Evolution on true-match rate.}
		%\footnotesize{(a) Evolution on association rate.} &
		%\footnotesize{(b) Evolution on true-match rate.}
	\end{tabular}
	\vspace{0.5em}
	\caption{Evolution on cross-camera tracklet association. The shaded areas denote the varying range of 10-split results repeated on PRID 2011 and iLIDS-VID. Best viewed in colour.}
	\vspace{-1em}
	\label{fig:ablation}
\end{figure}
To understand the effect of utilising the cyclic ranking consistency 
to discover highly associated tracklets during training, 
we track the proportion of {\em cross-camera anchors} that 
are updated to denote the cross-camera representation 
by merging two highly associated tracklets ({\em intra-camera anchors}). 
%conditioned on the {\em cyclic ranking consistency} 
%(Eq.~\eqref{eq:cyclic}, \eqref{eq:a_anchor_update}) during training. 
%To understand the effect of applying cycle-consistency in model
%training, we tracked the proportion 
%of tracklets which are cyclic consistent cross-camera matches
%(Eq.~\eqref{eq:cyclic}, \eqref{eq:a_anchor_update}) in the course of model
%training. 
%
Figure~\ref{fig:ablation}(a) shows that on PRID 2011 and iLIDS-VID, 
90+\% tracklets find their highly associated tracklets under another camera at the end of training. 
On the much noisier large-scale MARS dataset, 
the DAL can still associate more than half of tracklets (>50\%) across cameras. 
Importantly, as seen in Figure~\ref{fig:ablation}(b), among
self-discovered associated cross-camera tracklet pairs, the percentage
of true-match pairs at the end of training  
is approximately 90\% on PRID 2011, 75\% on iLIDS-VID, and 77\% on
MARS, respectively.
%Importantly, among self-discovered cross-camera associate tracklet pairs, the percentage of true matching pairs
%is 96.8\% on PRID 2011, 97.5\% on iLIDS-VID, and 95.1\% on MARS, respectively.
% as validated by the ground-truth ID labels.
%
This shows compellingly the strong capability of 
DAL in self-discovering the unknown cross-camera tracklet associations 
without learning from manually labelled data. %, which explicitly shows the performance 

\begin{table}[!t]
	\centering
	\footnotesize
	\setlength{\tabcolsep}{0.5em}
	\begin{tabular}{l|cccc|cccc|cccc|c}
		\hlineB{2}
		Datasets & \multicolumn{4}{c|}{PRID 2011} & \multicolumn{4}{c|}{iLIDS-VID} & \multicolumn{5}{c}{MARS} \\
		\hline
		Rank@$k$ & 1 & 5 & 10 & 20 & 1 & 5 & 10 & 20 & 1 & 5 & 10 & 20 & mAP \\ \hline\hline
		%DAL (w $\mathcal{L}_{I}{+}\mathcal{L}_{C}$) 
		DAL ($\mathcal{L}_{I} {+} \mathcal{L}_{C}$)
		& 
		{\color{red} 84.6} & {96.3} & {98.4} & {99.1} &
		%{\color{red} 86.1} & 96.9 & 98.3 & 99.2 & 
		{\color{red} 52.8} & {\color{red} 76.7} & 83.4 & {\color{red} 91.6} & 
		49.3 & 65.9 & 72.2 & 77.9 & 23.0
		\\
		ID-Supervised & 
		84.3 & {\color{red} 98.1} & {\color{red} 99.2} & {\color{red} 99.8} & 
		51.5 & 76.0 & {\color{red} 83.8} & 89.9 & 
		{\color{red} 71.8} & {\color{red} 86.8} & {\color{red} 90.7} & {\color{red} 93.3} & {\color{red} 51.5} \\
		\hlineB{2}
	\end{tabular}
	\vspace{1.0em}
	\caption{Comparison with supervised counterparts. {\color{red} Red}: the best performance. CNN: MobileNet.}
	\label{tab:counterpart}
	\vspace{-1em}
\end{table}

\vspace{0.1em}
\noindent {\bf Comparison with supervised counterparts.} 
We further compare DAL against the supervised counterpart 
trained using ID labelled data with the identical CNN
architecture (MobileNet), denoted as ID-Supervised.
This ID-Supervised is trained by the cross-entropy loss computed on
the ID labels. Results in Table~\ref{tab:counterpart} show that:
(1) On PRID 2011 and iLIDS-VID, DAL performs similarly well as 
the ID-Supervised. 
This is highly consistent with our observations of high tracklet association rate in in Figure~\ref{fig:ablation}, 
%both high tracklet association rate and true-match rate in Figure~\ref{fig:ablation}, 
%indicating that effectively discovering most cross-camera tracklet associations correctly 
indicating that discovering more cross-camera highly associated tracklets 
can help to learn a more discriminative re-id model that is robust to cross-camera variations. 
%which show most cross-camera tracklet associations are correctly 
%%such that most correct cross-camera tracklet associations are
%discovered by DAL to effectively learn from cross-camera variations. 
%
(2) On MARS, there is a clear performance gap between
the supervised and unsupervised models.
%
%This is largely due to the more challenging re-id task arising from discovering 
This is largely due to a relatively low tracklet association rate arising from the difficulty of discovering
cross-camera tracklet associations in a larger identity population among much 
noisier tracklets, as indicated in Figure~\ref{fig:ablation}(a).

%-------------------------------------------------------------------------
\vspace{-0.2cm}
\section{Conclusions}
In this work, we present a novel {\em Deep Association Learning}
(DAL) scheme for unsupervised video person re-id using  
unlabelled video tracklets extracted from surveillance video data. 
Our DAL permits deep re-id models to be trained without any ID labelling
for training data, which is therefore more scalable to deployment
on large-sized surveillance video data than supervised learning based models. 
In contrast to existing unsupervised video re-id methods that either require more
stringent one-camera ID labelling or per-camera tracklet
filtering, DAL is capable of learning to
automatically discover the more reliable cross-camera tracklet associations
for addressing the video re-id task without utilising ID labels. %stringent requirement. %such special assumptions.
This is achieved by jointly optimising two margin-based association
losses formulated based on the 
{\em local space-time consistency} and {\em global cyclic ranking consistency}. % between tracklets in different camera views.
Extensive comparative experiments on three video person
re-id benchmarks show compellingly the clear advantages of the proposed
DAL scheme over a wide variety of state-of-the-art unsupervised video person re-id methods.
We also provide detailed component analyses to further discuss the insights on how
each part of our method design contributes towards the overall model performance.

\vspace{-0.2cm}
\section*{Acknowledgements}
%{\bf Acknowledgements}
{This work was partly supported by the China Scholarship Council, Vision Semantics Limited, the Royal Society Newton Advanced Fellowship Programme (NA150459), and Innovate UK Industrial Challenge Project on Developing and Commercialising Intelligent Video Analytics Solutions for Public Safety (98111-571149).}

%-------------------------------------------------------------------------
\newpage
\bibliography{reference}

\begin{thebibliography}{52}
\providecommand{\natexlab}[1]{#1}
\providecommand{\url}[1]{\texttt{#1}}
\expandafter\ifx\csname urlstyle\endcsname\relax
  \providecommand{\doi}[1]{doi: #1}\else
  \providecommand{\doi}{doi: \begingroup \urlstyle{rm}\Url}\fi

\bibitem[Abadi et~al.(2016)Abadi, Barham, Chen, Chen, Davis, Dean, Devin,
  Ghemawat, Irving, Isard, et~al.]{abadi2016tensorflow}
Mart{\'\i}n Abadi, Paul Barham, Jianmin Chen, Zhifeng Chen, Andy Davis, Jeffrey
  Dean, Matthieu Devin, Sanjay Ghemawat, Geoffrey Irving, Michael Isard, et~al.
\newblock Tensorflow: A system for large-scale machine learning.
\newblock In \emph{OSDI}, volume~16, pages 265--283, 2016.

\bibitem[Ahmed et~al.(2015)Ahmed, Jones, and Marks]{ahmed2015improved}
Ejaz Ahmed, Michael Jones, and Tim~K Marks.
\newblock An improved deep learning architecture for person re-identification.
\newblock In \emph{IEEE Conference on Computer Vision and Pattern Recognition},
  2015.

\bibitem[Bengio et~al.(2013)Bengio, Courville, and
  Vincent]{bengio2013representation}
Yoshua Bengio, Aaron Courville, and Pascal Vincent.
\newblock Representation learning: A review and new perspectives.
\newblock \emph{IEEE Transactions on Pattern Analysis and Machine
  Intelligence}, 2013.

\bibitem[Chen et~al.(2016)Chen, Guo, and Lai]{chen2016deep}
Shi-Zhe Chen, Chun-Chao Guo, and Jian-Huang Lai.
\newblock Deep ranking for person re-identification via joint representation
  learning.
\newblock \emph{IEEE Transactions on Image Processing}, 2016.

\bibitem[Chen et~al.(2017{\natexlab{a}})Chen, Chen, Zhang, and
  Huang]{chen2017beyond}
Weihua Chen, Xiaotang Chen, Jianguo Zhang, and Kaiqi Huang.
\newblock Beyond triplet loss: a deep quadruplet network for person
  re-identification.
\newblock In \emph{IEEE Conference on Computer Vision and Pattern Recognition},
  2017{\natexlab{a}}.

\bibitem[Chen et~al.(2017{\natexlab{b}})Chen, Zhu, Gong,
  et~al.]{chen2017person}
Yanbei Chen, Xiatian Zhu, Shaogang Gong, et~al.
\newblock Person re-identification by deep learning multi-scale
  representations.
\newblock In \emph{Workshop of IEEE International Conference on Computer
  Vision}, 2017{\natexlab{b}}.

\bibitem[Cheng et~al.(2016)Cheng, Gong, Zhou, Wang, and Zheng]{cheng2016person}
De~Cheng, Yihong Gong, Sanping Zhou, Jinjun Wang, and Nanning Zheng.
\newblock Person re-identification by multi-channel parts-based cnn with
  improved triplet loss function.
\newblock In \emph{IEEE Conference on Computer Vision and Pattern Recognition},
  2016.

\bibitem[Dehghan et~al.(2015)Dehghan, Modiri~Assari, and
  Shah]{dehghan2015gmmcp}
Afshin Dehghan, Shayan Modiri~Assari, and Mubarak Shah.
\newblock Gmmcp tracker: Globally optimal generalized maximum multi clique
  problem for multiple object tracking.
\newblock In \emph{IEEE Conference on Computer Vision and Pattern Recognition},
  2015.

\bibitem[Deng et~al.(2009)Deng, Dong, Socher, Li, Li, and
  Fei-Fei]{deng2009imagenet}
Jia Deng, Wei Dong, Richard Socher, Li-Jia Li, Kai Li, and Li~Fei-Fei.
\newblock Imagenet: A large-scale hierarchical image database.
\newblock In \emph{IEEE Conference on Computer Vision and Pattern Recognition},
  2009.

\bibitem[Ding et~al.(2015)Ding, Lin, Wang, and Chao]{ding2015deep}
Shengyong Ding, Liang Lin, Guangrun Wang, and Hongyang Chao.
\newblock Deep feature learning with relative distance comparison for person
  re-identification.
\newblock \emph{Pattern Recognition}, 2015.

\bibitem[Felzenszwalb et~al.(2010)Felzenszwalb, Girshick, McAllester, and
  Ramanan]{felzenszwalb2010object}
Pedro~F Felzenszwalb, Ross~B Girshick, David McAllester, and Deva Ramanan.
\newblock Object detection with discriminatively trained part-based models.
\newblock \emph{IEEE Transactions on Pattern Analysis and Machine
  Intelligence}, 2010.

\bibitem[Godard et~al.(2017)Godard, Mac~Aodha, and
  Brostow]{godard2017unsupervised}
Cl{\'e}ment Godard, Oisin Mac~Aodha, and Gabriel~J Brostow.
\newblock Unsupervised monocular depth estimation with left-right consistency.
\newblock In \emph{IEEE Conference on Computer Vision and Pattern Recognition},
  2017.

\bibitem[Gong et~al.(2014)Gong, Cristani, Yan, and Loy]{gong2014person}
Shaogang Gong, Marco Cristani, Shuicheng Yan, and Chen~Change Loy.
\newblock \emph{Person re-identification}.
\newblock Springer, 2014.

\bibitem[He et~al.(2016)He, Zhang, Ren, and Sun]{he2016deep}
Kaiming He, Xiangyu Zhang, Shaoqing Ren, and Jian Sun.
\newblock Deep residual learning for image recognition.
\newblock In \emph{IEEE Conference on Computer Vision and Pattern Recognition},
  2016.

\bibitem[Hermans et~al.(2017)Hermans, Beyer, and Leibe]{hermans2017defense}
Alexander Hermans, Lucas Beyer, and Bastian Leibe.
\newblock In defense of the triplet loss for person re-identification.
\newblock \emph{arXiv preprint arXiv:1703.07737}, 2017.

\bibitem[Hirzer et~al.(2011)Hirzer, Beleznai, Roth, and
  Bischof]{hirzer2011person}
Martin Hirzer, Csaba Beleznai, Peter~M Roth, and Horst Bischof.
\newblock Person re-identification by descriptive and discriminative
  classification.
\newblock In \emph{Scandinavian Conference on Image Analysis}, 2011.

\bibitem[Howard et~al.(2017)Howard, Zhu, Chen, Kalenichenko, Wang, Weyand,
  Andreetto, and Adam]{howard2017mobilenets}
Andrew~G Howard, Menglong Zhu, Bo~Chen, Dmitry Kalenichenko, Weijun Wang,
  Tobias Weyand, Marco Andreetto, and Hartwig Adam.
\newblock Mobilenets: Efficient convolutional neural networks for mobile vision
  applications.
\newblock \emph{arXiv preprint arXiv:1704.04861}, 2017.

\bibitem[Karanam et~al.(2015)Karanam, Li, and Radke]{karanam2015person}
Srikrishna Karanam, Yang Li, and Richard~J Radke.
\newblock Person re-identification with discriminatively trained viewpoint
  invariant dictionaries.
\newblock In \emph{IEEE International Conference on Computer Vision}, 2015.

\bibitem[Khan and Bremond(2016)]{khan2016unsupervised}
Furqan~M Khan and Francois Bremond.
\newblock Unsupervised data association for metric learning in the context of
  multi-shot person re-identification.
\newblock In \emph{IEEE International Conference on Advanced Video and Signal
  Based Surveillance}, 2016.

\bibitem[Li et~al.(2018{\natexlab{a}})Li, Bak, Carr, and Wang]{li2018diversity}
Shuang Li, Slawomir Bak, Peter Carr, and Xiaogang Wang.
\newblock Diversity regularized spatiotemporal attention for video-based person
  re-identification.
\newblock In \emph{IEEE Conference on Computer Vision and Pattern Recognition},
  2018{\natexlab{a}}.

\bibitem[Li et~al.(2014)Li, Zhao, Xiao, and Wang]{li2014deepreid}
Wei Li, Rui Zhao, Tong Xiao, and Xiaogang Wang.
\newblock Deepreid: Deep filter pairing neural network for person
  re-identification.
\newblock In \emph{IEEE Conference on Computer Vision and Pattern Recognition},
  2014.

\bibitem[Li et~al.(2017)Li, Zhu, and Gong]{li2017person}
Wei Li, Xiatian Zhu, and Shaogang Gong.
\newblock Person re-identification by deep joint learning of multi-loss
  classification.
\newblock In \emph{International Joint Conference of Artificial Intelligence},
  2017.

\bibitem[Li et~al.(2018{\natexlab{b}})Li, Zhu, and Gong]{li2018harmonious}
Wei Li, Xiatian Zhu, and Shaogang Gong.
\newblock Harmonious attention network for person re-identification.
\newblock In \emph{IEEE Conference on Computer Vision and Pattern Recognition},
  2018{\natexlab{b}}.

\bibitem[Liu et~al.(2016)Liu, Zha, Tian, Liu, Yao, Ling, and Mei]{liu2016multi}
Jiawei Liu, Zheng-Jun Zha, QI~Tian, Dong Liu, Ting Yao, Qiang Ling, and Tao
  Mei.
\newblock Multi-scale triplet cnn for person re-identification.
\newblock In \emph{ACM International Conference on Multimedia}, 2016.

\bibitem[Liu et~al.(2015)Liu, Ma, Zhang, and Huang]{liu2015spatio}
Kan Liu, Bingpeng Ma, Wei Zhang, and Rui Huang.
\newblock A spatio-temporal appearance representation for video-based
  pedestrian re-identification.
\newblock In \emph{IEEE International Conference on Computer Vision}, 2015.

\bibitem[Liu et~al.(2017)Liu, Wang, and Lu]{liu2017stepwise}
Zimo Liu, Dong Wang, and Huchuan Lu.
\newblock Stepwise metric promotion for unsupervised video person
  re-identification.
\newblock In \emph{IEEE International Conference on Computer Vision}, 2017.

\bibitem[Ma et~al.(2017)Ma, Zhu, Gong, Xie, Hu, Lam, and Zhong]{ma2017person}
Xiaolong Ma, Xiatian Zhu, Shaogang Gong, Xudong Xie, Jianming Hu, Kin-Man Lam,
  and Yisheng Zhong.
\newblock Person re-identification by unsupervised video matching.
\newblock \emph{Pattern Recognition}, 2017.

\bibitem[McLaughlin et~al.(2016)McLaughlin, del Rincon, and
  Miller]{mclaughlin2016recurrent}
Niall McLaughlin, Jesus~Martinez del Rincon, and Paul Miller.
\newblock Recurrent convolutional network for video-based person
  re-identification.
\newblock In \emph{IEEE Conference on Computer Vision and Pattern Recognition},
  2016.

\bibitem[Paisitkriangkrai et~al.(2015)Paisitkriangkrai, Shen, and van~den
  Hengel]{paisitkriangkraicvpr15}
Sakrapee Paisitkriangkrai, Chunhua Shen, and Anton van~den Hengel.
\newblock Learning to rank in person re-identification with metric ensembles.
\newblock In \emph{IEEE Conference on Computer Vision and Pattern Recognition},
  Boston, Massachusetts, USA, 2015.

\bibitem[Prosser et~al.(2010)Prosser, Zheng, Gong, and Xiang]{prosserbmvc10}
B.~Prosser, W-S. Zheng, S.~Gong, and T.~Xiang.
\newblock Person re-identification by support vector ranking.
\newblock In \emph{British Machine Vision Conference}, Aberystwyth, Wales,
  September 2010.

\bibitem[Sener et~al.(2016)Sener, Song, Saxena, and
  Savarese]{sener2016learning}
Ozan Sener, Hyun~Oh Song, Ashutosh Saxena, and Silvio Savarese.
\newblock Learning transferrable representations for unsupervised domain
  adaptation.
\newblock In \emph{Advances in Neural Information Processing Systems}, 2016.

\bibitem[Sun et~al.(2017)Sun, Zheng, Deng, and Wang]{sun2017svdnet}
Yifan Sun, Liang Zheng, Weijian Deng, and Shengjin Wang.
\newblock Svdnet for pedestrian retrieval.
\newblock In \emph{IEEE International Conference on Computer Vision}, 2017.

\bibitem[Tieleman and Hinton(2012)]{tieleman2012lecture}
Tijmen Tieleman and Geoffrey Hinton.
\newblock Lecture 6.5-rmsprop: Divide the gradient by a running average of its
  recent magnitude.
\newblock \emph{COURSERA: Neural networks for machine learning}, pages 26--31,
  2012.

\bibitem[Wang et~al.(2016{\natexlab{a}})Wang, Zuo, Lin, Zhang, and
  Zhang]{wang2016joint}
Faqiang Wang, Wangmeng Zuo, Liang Lin, David Zhang, and Lei Zhang.
\newblock Joint learning of single-image and cross-image representations for
  person re-identification.
\newblock In \emph{IEEE Conference on Computer Vision and Pattern Recognition},
  2016{\natexlab{a}}.

\bibitem[Wang et~al.(2018{\natexlab{a}})Wang, Zhu, Gong, and
  Xiang]{wang2018person}
Hanxiao Wang, Xiatian Zhu, Shaogang Gong, and Tao Xiang.
\newblock Person re-identification in identity regression space.
\newblock \emph{International Journal of Computer Vision}, 2018{\natexlab{a}}.

\bibitem[Wang et~al.(2018{\natexlab{b}})Wang, Zhu, Gong, and Li]{wang2018reid}
Jingya Wang, Xiatian Zhu, Shaogang Gong, and Wei Li.
\newblock Transferable joint attribute-identity deep learning for unsupervised
  person re-identification.
\newblock In \emph{IEEE Conference on Computer Vision and Pattern Recognition},
  2018{\natexlab{b}}.

\bibitem[Wang et~al.(2014)Wang, Gong, Zhu, and Wang]{wang2014person}
Taiqing Wang, Shaogang Gong, Xiatian Zhu, and Shengjin Wang.
\newblock Person re-identification by video ranking.
\newblock In \emph{European Conference on Computer Vision}, 2014.

\bibitem[Wang et~al.(2016{\natexlab{b}})Wang, Gong, Zhu, and
  Wang]{wang2016person}
Taiqing Wang, Shaogang Gong, Xiatian Zhu, and Shengjin Wang.
\newblock Person re-identification by discriminative selection in video
  ranking.
\newblock \emph{IEEE Transactions on Pattern Analysis and Machine
  Intelligence}, 2016{\natexlab{b}}.

\bibitem[Xiao et~al.(2016)Xiao, Li, Ouyang, and Wang]{xiao2016learning}
Tong Xiao, Hongsheng Li, Wanli Ouyang, and Xiaogang Wang.
\newblock Learning deep feature representations with domain guided dropout for
  person re-identification.
\newblock In \emph{IEEE Conference on Computer Vision and Pattern Recognition}.
  IEEE, 2016.

\bibitem[Xu et~al.(2017)Xu, Cheng, Gu, Yang, Chang, and Zhou]{xu2017jointly}
Shuangjie Xu, Yu~Cheng, Kang Gu, Yang Yang, Shiyu Chang, and Pan Zhou.
\newblock Jointly attentive spatial-temporal pooling networks for video-based
  person re-identification.
\newblock In \emph{IEEE Conference on Computer Vision and Pattern Recognition},
  2017.

\bibitem[Yan et~al.(2016)Yan, Ni, Song, Ma, Yan, and Yang]{yan2016person}
Yichao Yan, Bingbing Ni, Zhichao Song, Chao Ma, Yan Yan, and Xiaokang Yang.
\newblock Person re-identification via recurrent feature aggregation.
\newblock In \emph{European Conference on Computer Vision}, 2016.

\bibitem[Ye et~al.(2017)Ye, Ma, Zheng, Li, and Yuen]{ye2017dynamic}
Mang Ye, Andy~J Ma, Liang Zheng, Jiawei Li, and Pong~C Yuen.
\newblock Dynamic label graph matching for unsupervised video
  re-identification.
\newblock In \emph{IEEE International Conference on Computer Vision}, 2017.

\bibitem[Yi et~al.(2014)Yi, Lei, Liao, and Li]{yi2014deep}
Dong Yi, Zhen Lei, Shengcai Liao, and Stan~Z Li.
\newblock Deep metric learning for person re-identification.
\newblock In \emph{IEEE International Conference on Pattern Recognition}, 2014.

\bibitem[You et~al.(2016)You, Wu, Li, and Zheng]{you2016top}
Jinjie You, Ancong Wu, Xiang Li, and Wei-Shi Zheng.
\newblock Top-push video-based person re-identification.
\newblock In \emph{IEEE Conference on Computer Vision and Pattern Recognition},
  2016.

\bibitem[Zheng et~al.(2016)Zheng, Bie, Sun, Wang, Su, Wang, and
  Tian]{zheng2016mars}
Liang Zheng, Zhi Bie, Yifan Sun, Jingdong Wang, Chi Su, Shengjin Wang, and
  Qi~Tian.
\newblock Mars: A video benchmark for large-scale person re-identification.
\newblock In \emph{European Conference on Computer Vision}, 2016.

\bibitem[Zheng et~al.(2013)Zheng, Gong, and Xiang]{zhengpami13}
W-S. Zheng, S.~Gong, and T.~Xiang.
\newblock Re-identification by relative distance comparison.
\newblock \emph{IEEE Transactions on Pattern Analysis and Machine
  Intelligence}, 2013.

\bibitem[Zheng et~al.(2017)Zheng, Zheng, and Yang]{zheng2017unlabeled}
Zhedong Zheng, Liang Zheng, and Yi~Yang.
\newblock Unlabeled samples generated by gan improve the person
  re-identification baseline in vitro.
\newblock In \emph{IEEE International Conference on Computer Vision}, 2017.

\bibitem[Zhong et~al.(2018)Zhong, Zheng, Zheng, Li, and Yang]{zhong2017camera}
Zhun Zhong, Liang Zheng, Zhedong Zheng, Shaozi Li, and Yi~Yang.
\newblock Camera style adaptation for person re-identification.
\newblock In \emph{IEEE Conference on Computer Vision and Pattern Recognition},
  2018.

\bibitem[Zhou et~al.(2016)Zhou, Krahenbuhl, Aubry, Huang, and
  Efros]{zhou2016learning}
Tinghui Zhou, Philipp Krahenbuhl, Mathieu Aubry, Qixing Huang, and Alexei~A
  Efros.
\newblock Learning dense correspondence via 3d-guided cycle consistency.
\newblock In \emph{IEEE Conference on Computer Vision and Pattern Recognition},
  2016.

\bibitem[Zhou et~al.(2017)Zhou, Huang, Wang, Wang, and Tan]{zhou2017see}
Zhen Zhou, Yan Huang, Wei Wang, Liang Wang, and Tieniu Tan.
\newblock See the forest for the trees: Joint spatial and temporal recurrent
  neural networks for video-based person re-identification.
\newblock In \emph{IEEE Conference on Computer Vision and Pattern Recognition},
  2017.

\bibitem[Zhu et~al.(2016)Zhu, Jing, Wu, and Feng]{zhu2016video}
Xiaoke Zhu, Xiao-Yuan Jing, Fei Wu, and Hui Feng.
\newblock Video-based person re-identification by simultaneously learning
  intra-video and inter-video distance metrics.
\newblock In \emph{International Joint Conference of Artificial Intelligence},
  2016.

\bibitem[Zhu et~al.(2017)Zhu, Wu, Huang, and Zheng]{zhu2017fast}
Xiatian Zhu, Botong Wu, Dongcheng Huang, and Wei-Shi Zheng.
\newblock Fast openworld person re-identification.
\newblock \emph{IEEE Transactions on Image Processing}, 2017.

\end{thebibliography}

\end{document}